%% file: ms.tex
\pdfoutput=1
\documentclass[11pt,a4paper]{article}
\usepackage[dvipsnames]{xcolor}
\usepackage[hyperref]{emnlp2020}
\usepackage{times}
\usepackage{latexsym}

\usepackage{microtype}

\usepackage{amsfonts}
\usepackage{amsmath}
\usepackage{comment}
\usepackage{url}
\usepackage{graphicx}
\usepackage{threeparttable}
\usepackage{svg}
\usepackage{multirow}

\aclfinalcopy 
\def\aclpaperid{1093} 

\input{macros}

\title{\papertitle{}}

\author{Ethan C. Chau$^{\dagger \Diamond}$ \quad
  Lucy H. Lin$^\dagger$ \quad
  Noah A. Smith$^{\dagger \star}$ \\
  $^\dagger$Paul G. Allen School of Computer Science \& Engineering, University of Washington \\
  $^\Diamond$Department of Linguistics, University of Washington \\
  $^\star$Allen Institute for Artificial Intelligence \\
  {\tt \{echau18,lucylin,nasmith\}@cs.washington.edu}}

\date{}

\begin{document}
\maketitle
\begin{abstract}
    Pretrained multilingual contextual representations have shown great success, but due to the limits of their pretraining data, their benefits do not apply equally to all language varieties.  This presents a challenge for language varieties unfamiliar to these models, whose labeled \emph{and unlabeled} data is too limited to train a monolingual model effectively.  We propose the use of additional language-specific pretraining and vocabulary augmentation to adapt multilingual models to low-resource settings.  Using dependency parsing of four diverse low-resource language varieties as a case study, we show that these methods significantly improve performance over baselines, especially in the lowest-resource cases, and demonstrate the importance of the relationship between such models' pretraining data and target language varieties.
\end{abstract}

\input{sections/1_intro}
\input{sections/2_method}
\input{sections/3_evaluation}
\input{sections/4_results}
\input{sections/5_relatedwork}
\input{sections/6_conclusion}
\input{sections/a_acknowledgments}

\bibliography{ms}
\bibliographystyle{acl_natbib}

\newpage
\appendix
\input{sections/a_appendix}

\end{document}

%% file: macros.tex
\newcommand{\err}[1]{\small{$\pm$ #1}}
\newcommand{\papertitle}{Parsing with Multilingual BERT, a Small Corpus, and a Small Treebank}

\newcommand{\ga}{\textsc{ga}}
\newcommand{\mt}{\textsc{mt}}
\newcommand{\sing}{\textsc{sing}}
\newcommand{\vi}{\textsc{vi}}

\newcommand{\fttrained}{\textsc{fastT}}
\newcommand{\elmotrained}{\textsc{elmo}}
\newcommand{\mbert}{\textsc{mBert}}

\newcommand{\pt}{\textsc{lapt}}
\newcommand{\va}{\textsc{va}}
\newcommand{\tva}{\textsc{tva}}
\newcommand{\ft}{\textsc{ft}}

%% file: sections/1_intro.tex
\section{Introduction}

Contextual word representations (CWRs) from pretrained language models have improved many NLP systems.  Such language models include BERT \citep{devlin-etal-2019-bert} and ELMo \citep{peters-etal-2018-deep}, which are conventionally ``pretrained'' on large unlabeled datasets before their internal representations are ``finetuned'' during supervised training on downstream tasks like parsing.  However, many language varieties\footnote{Sociolinguists define ``language varieties'' broadly to encompass any distinct form of a language.  In addition to standard varieties (conventionally referred to as ``languages''), this includes dialects, registers, and styles \citep{trudgill-glossary-10.3366/j.ctvxcrtfv.4}.} lack large annotated and even unannotated datasets, raising questions about the broad applicability of such data-hungry methods.

One exciting way to compensate for the lack of unlabeled data in low-resource language varieties is to finetune a large, \emph{multilingual} language model that has been pretrained on the union of many languages' data \citep{devlin-etal-2019-bert,lample2019cross}.  This enables the model to transfer some of what it learns from high-resource languages to low-resource ones, demonstrating benefits over monolingual methods in some cases \citep{conneau2019unsupervised,tsai-etal-2019-small}, though not always \citep{agerri2020text,ronnqvist-etal-2019-multilingual}.

Specifically, multilingual models face the transfer-dilution tradeoff \citep{conneau2019unsupervised}: increasing the number of languages during pretraining improves positive crosslingual transfer but decreases the model capacity allocated to each language.  Furthermore, such models are only pretrained on a finite amount of data and may lack exposure to specialized domains of certain languages or even entire low-resource language varieties.  The result is a challenge for these language varieties, which must rely on positive transfer from a sufficient number of similar high-resource languages.  Indeed, \citet{wu-dredze-2020-languages} find that multilingual models often underperform monolingual baselines for such languages and question their off-the-shelf viability.

We take inspiration from previous work on domain adaptation, where general-purpose monolingual models have been effectively adapted to specialized domains through additional pretraining on domain-specific corpora \citep{dontstoppretraining2020}. We hypothesize that we can improve the performance of multilingual models on low-resource language varieties analogously, through additional pretraining on \emph{language}-specific corpora.

However, additional pretraining on more data in the target language does not ensure its full representation in the model's vocabulary, which is constructed to maximally represent the model's original pretraining data \citep{sennrich-etal-2016-neural,wu2016googles}. 
\citet{artetxe-etal-2020-cross} find that target languages' representation in the vocabulary affects these models' transferability,
suggesting that language varieties on the fringes of the vocabulary may not be sufficiently well-modeled.  Can we incorporate vocabulary from the target language into multilingual models' existing alignment?

We introduce the use of additional language-specific pretraining for multilingual CWRs in a low-resource setting, \emph{before} use in a downstream task; to better model language-specific tokens, we also augment the existing vocabulary with frequent tokens from the low-resource language (\S \ref{sec:method}).  Our experiments consider dependency parsing in four typologically diverse low-resource language varieties with different degrees of relatedness to a multilingual model's pretraining data (\S \ref{sec:experiments}).  Our results show that these methods consistently improve performance on each target variety, especially in the lowest-resource cases (\S \ref{sec:results}).  In doing so, we demonstrate the importance of accounting for the relationship between a multilingual model's pretraining data and the target language variety.  

Because the pretraining-finetuning paradigm is now ubiquitous, many experimental findings for one task can now inform work on other tasks.  Thus, our findings on dependency parsing---whose annotated datasets cover many more low-resource language varieties than those of other NLP tasks---are expected to interest researchers and practitioners facing low-resource situations for other tasks.  To this end, we make our code, data, and hyperparameters publicly available.\footnote{\label{fn:code}\url{https://github.com/ethch18/parsing-mbert}}

%% file: sections/2_method.tex
\section{Overview}\label{sec:method}

We are chiefly concerned with the adaptation of pretrained multilingual models to a target language by optimally using available data.  As a case study, we use the multilingual cased BERT model (\mbert{}) of \citet{devlin-etal-2019-bert}, a transformer-based \citep{Vaswani2017AttentionIA} language model which has produced strong CWRs for many languages \citep[\emph{inter alia}]{kondratyuk-straka-2019-75}.  \mbert{} is pretrained on the 104 languages with the most Wikipedia data and encodes input tokens using a fixed wordpiece vocabulary \citep{wu2016googles} learned from this data.  Low-resource languages are slightly oversampled in its pretraining data, but high resource languages are still more prevalent, resulting in a language imbalance.\footnote{Sampling is done based on an exponentially smoothed distribution of the amount of data in each language, which slightly increases the representation of low-resource languages.  See \url{https://github.com/google-research/bert/blob/master/multilingual.md} for more details.}

We observe that two types of target language varieties may be disadvantaged by this training scheme: the lowest-resource languages in \mbert{}'s pretraining data (which we call Type 1); and unseen low-resource languages (Type 2).  Although Type 1 languages are oversampled during training, they are still overshadowed by high-resource languages.  Type 2 languages must rely purely on crosslingual vocabulary overlap.  In both cases, the wordpieces that encode the input tokens in these languages may not fully capture the senses in which they are used, or they may be completely unseen.\footnote{Wordpiece tokenization is done greedily based on a fixed vocabulary.  The model returns a special ``unknown'' token for unseen characters and other subword units that cannot be represented by the vocabulary.}  However, other low-resource varieties with more representation in \mbert{}'s pretraining data (Type 0) may not be as disadvantaged.  Optimally using \mbert{} in low-resource settings thus requires accounting for limitations with respect to a target language variety.

\subsection{Methods}

We evaluate three methods of adapting \mbert{} to better model target language varieties.

\begin{table*}[t]
    \centering
    \begin{tabular}{lcrrrr}
        \hline
        \textbf{Language} & \textbf{Type} & \textbf{\# Sentences} & \textbf{\# Tokens} & \textbf{WP/Token} & \textbf{UNK Tokens} \\
        \hline
        \ga{} & 1 & 199k & 3.6M & 2.10 & 12807 \\
        \mt{} & 2 & 62k & 1M & 2.95 & 49791 \\
        \sing{} & 0 & 80k & 1.2M & 1.24 & 3 \\
        \vi{} & 0 & 255k & 5.6M & 1.33 & 6955 \\
        \hline
    \end{tabular}
    \caption{Unlabeled dataset statistics: number of sentences, number of tokens, average wordpieces per token, and tokens containing an unknown wordpiece under original \mbert{} vocabulary.  
    }
    \label{tab:data_summary}
\end{table*}

\paragraph{Language-Adaptive Pretraining (\pt{})}

Under the assumption that language varieties function analagously to domains for \mbert{}, we adapt the \emph{domain-adaptive pretraining} method of \citet{dontstoppretraining2020} to a multilingual setting.  With \emph{language-adaptive pretraining}, \mbert{} is pretrained for additional epochs on monolingual data in the target language variety to improve the alignment of the wordpiece embeddings.

\paragraph{Vocabulary Augmentation (\va{})}

To better model unseen or language-specific wordpieces, we explore performing \pt{} after augmenting \mbert{}'s vocabulary from a target language variety.  We train a new wordpiece vocabulary on monolingual data in the target language, tokenize the monolingual data with the new vocabulary, and augment \mbert{}'s vocabulary with the 99 most common wordpieces\footnote{\mbert{}'s fixed-size vocabulary contains 99 tokens designated as ``unused,'' whose representations were not updated during initial pretraining and can be repurposed for vocabulary augmentation without modifying the pretrained model.} in the new vocabulary that replaced the ``unknown'' wordpiece token.
Full details of this process are given in the Appendix.

\paragraph{Tiered Vocabulary Augmentation (\tva{})}

We consider a variant of \va{} with a larger learning rate for the embeddings of the 99 new wordpieces than for the other parameters, to explore these embeddings' potential to be learned more thoroughly without overfitting the model's remaining parameters.  Learning rate details are given in
the Appendix.

\subsection{Evaluation}

We perform evaluation on dependency parsing.  Following \citet{kondratyuk-straka-2019-75}, we take a weighted sum of the activations at each \mbert{} layer as the CWR for each token.  We then pass the representations into the graph-based dependency parser of \citet{dozat2016deep}. This parser, which is also used in related work \citep{kondratyuk-straka-2019-75,mulcaire-etal-2019-low,schuster2019crosslingual}, uses a biaffine attention mechanism between word representations to score a parse tree.

%% file: sections/3_evaluation.tex
\section{Experiments}\label{sec:experiments}

We consider two variants of each \mbert{} method: one in which the pretrained CWRs are frozen; and one where they are further finetuned during parser training (\ft{}).  Following prior work involving these two variants \citep{Beltagy2019SciBERT}, \ft{} variants perform biaffine attention directly on the outputs of \mbert{} instead of first passing them through a BiLSTM, as in \citet{dozat2016deep}.

We perform additional pretraining for up to 20 epochs, selecting our final models based on average validation LAS downstream.
Full training details are given in the Appendix.
We report average scores and standard errors based on five random initializations.  Code and data are publicly available (see footnote~\ref{fn:code}).

\subsection{Languages and Datasets}

We perform experiments on four typologically diverse low-resource languages: Irish (\ga{}), Maltese (\mt{}), Vietnamese (\vi{}), and Singlish (Singapore Colloquial English; \sing{}).  Singlish is an English-based creole spoken in Singapore, which incorporates lexical and syntactic borrowings from other languages spoken in Singapore: Chinese, Malay, and Tamil.  \citet{wang-etal-2017-universal} provide an extended motivation for evaluating on Singlish.

These language varieties are examplars of the three types discussed in \S \ref{sec:method}.  \mbert{} is trained on the 104 largest Wikipedias, which includes Irish and Vietnamese but \emph{excludes} Maltese and Singlish.  However, the Irish Wikipedia is several orders of magnitude smaller than the full Vietnamese one.  So, we view Irish and Maltese as Type 1 and Type 2 language varieties, respectively.  Though Singlish lacks its own Wikipedia and is likely not included in \mbert{}'s pretraining data \emph{per se}, its component languages (English, Chinese, Malay, and Tamil) are all well-represented in the data.  We thus consider it to be a Type 0 variety along with Vietnamese.

\paragraph{Unlabeled Datasets}

\begin{table*}
    \centering
    \begin{tabular}{lrrrr}
        \hline
         \multirow{2}{*}{\textbf{Representations}} & \textbf{Irish (\ga{})} & \textbf{Maltese (\mt{})} & \textbf{Singlish (\sing{})} & \textbf{Vietnamese (\vi{})} \\
        & Type 1 & Type 2 & Type 0 & Type 0 \\
        \hline
        \fttrained{} & 65.36 \err{1.33} & 68.23 \err{0.61} & 66.42 \err{0.92} & 53.37 \err{0.95} \\
        \elmotrained{} & 68.25 \err{0.37} & 74.33 \err{0.53} & 68.63 \err{1.04} & 56.91 \err{0.41} \\
        \hline
        \mbert{} & 68.19 \err{0.43} & 67.06 \err{0.61} & 74.01 \err{0.39} & 62.96 \err{0.41} \\
        \pt{} & \textbf{73.03 \err{0.25}} & 78.51 \err{0.41} & \textbf{76.48 \err{0.63}} & \textbf{64.67 \err{0.22}} \\
        \va{} & 72.68 \err{0.47} & \textbf{79.88 \err{0.55}} & \textbf{76.71 \err{0.70}} & 64.28 \err{0.44} \\
        \tva{} & \textbf{72.95 \err{0.30}} & 79.32 \err{0.26} & \textbf{76.88 \err{0.62}} & 64.40 \err{0.45} \\
        \hline
        \mbert{} + \ft{} & 72.67 \err{0.22} & 76.74 \err{0.35} & 78.24 \err{0.52} & 66.13 \err{0.38} \\
        \pt{} + \ft{} & 75.45 \err{0.28} & 82.77 \err{0.24} & 79.30 \err{0.57} & \textbf{67.50 \err{0.25}} \\
        \va{} + \ft{} & \textbf{76.17 \err{0.08}} & \textbf{83.53 \err{0.21}} & \textbf{79.89 \err{0.46}} & \textbf{67.28 \err{0.38}} \\
        \tva{} + \ft{} & 75.95 \err{0.24} & 83.10 \err{0.26} & \textbf{80.10 \err{0.39}} & \textbf{67.35 \err{0.30}} \\
        \hline
    \end{tabular}
    \caption{Results (LAS) on downstream UD parsing, with standard deviations from five random initializations.  \textbf{Bolded} results are within one standard deviation of the maximum for each category (frozen/\ft{}).
    }
    \label{tab:dependency_parsing}
\end{table*}

Additional pretraining for Irish, Maltese, and Vietnamese uses unlabeled articles from Wikipedia.  To simulate a truly low-resource setting for Vietnamese, we use a random sample of 5\% of the articles.  Singlish data is crawled from the \textit{SG Talk Forum}\footnote{\url{https://sgTalk.com}} online forum and provided by \citet{wang-etal-2017-universal}.
To ensure robust evaluation, we remove all sentences that appear in the labeled validation and test sets from the unlabeled data.  Full details are provided in
the Appendix.

Tab.~\ref{tab:data_summary} gives the average number of wordpieces per token and the number of tokens with unknown wordpieces in each of the unlabeled datasets, computed based on the original \mbert{} vocabulary.  While the high number of wordpieces per token for Irish and Maltese may be due in part to morphological richness, it also suggests that these languages stand to benefit more from improved alignment of the wordpieces' embeddings.  Furthermore, the higher rates of unknown wordpieces leave room for enhanced performance with an improved vocabulary.

\paragraph{Labeled Datasets}

Parsers for Irish, Maltese, and Vietnamese are trained on the corresponding treebanks and train/test splits from Universal Dependencies 2.5 \citep{ud-dataset}: IDT, MUDT, and VTB, respectively.  For Singlish, we use the extended treebank component of \citet{wang-tallip}, which we randomly partition into train (80\%), valid. (10\%), and test sets (10\%).\footnote{Our partition of the data is available at \url{https://github.com/ethch18/parsing-mbert}.}  We use the provided gold word segmentation but no POS tag features.

\subsection{Baselines}\label{subsec:baselines}

For each language, we evaluate the performance of \mbert{} in frozen and \ft{} variants, without any adaptations.  We additionally benchmark each method against strong prior work that represents conventional approaches for representing low-resource languages: static fastText embeddings \citep[\fttrained{};][]{bojanowski-etal-2017-enriching}, which can be learned effectively even on small datasets; and monolingual ELMo models \citep[\elmotrained{};][]{peters-etal-2018-deep}, a monolingual contextual approach.  We choose ELMo over training a new BERT model because the high computational and data requirements of the latter make it unviable in a low-resource setting.  Both baselines are trained on our unlabeled datasets.

%% file: sections/4_results.tex
\section{Results and Discussion}\label{sec:results}

Tab.~\ref{tab:dependency_parsing} shows the performance of each of the method variants on the four Universal Dependencies datasets, with standard deviations from five different initializations.
Our experiments demonstrate that additional language-specific pretraining results in more effective representations.  \pt{} consistently outperforms baselines, especially for Irish and Maltese, where overlap with the original pretraining data is low and frozen \mbert{} underperforms \elmotrained{}.  This suggests that the insights of \citet{dontstoppretraining2020} on additional pretraining for domain adaptation are also applicable to transferring multilingual models to low-resource languages, even without much additional data.

\pt{} with our vocabulary augmentation methods yield small but significant improvements over \pt{} alone, especially for \ft{} configurations and Type 1/2 languages.  
This demonstrates that accurate vocabulary modeling is important for improving representations in the target language, and that \va{} is an effective methods for doing so while maintaining overall alignment.
However, \tva{} rarely outperforms \va{} significantly, suggesting that accelerated learning of the new embeddings does not benefit the model overall.

The relative error reductions between unadapted \mbert{} and each of our methods correlates with each language variety's relationship to \mbert{} pretraining data.  Maltese (Type 2) improves by up to 39\% and Irish (Type 1) by up to 15\%, compared to 11\% for Singlish and 5\% for Vietnamese (both Type 0).  While this trend is by no means comprehensive, it suggests that effective use of \mbert{} requires considering the target language variety.

Our results thus support our hypotheses and give insight to the limitations of \mbert{}.  Wordpieces appear in different contexts in different languages, and \mbert{} initially lacks enough exposure to wordpiece usage in Type 1/2 target languages to outperform baselines.  However, increased exposure through additional language-specific pretraining can ameliorate this issue.  Likewise, despite \mbert{}'s attempt to balance its pretraining data, the existing vocabulary still favors languages that have been seen more.  Augmenting the vocabulary can produce additional improvement for languages with greater proportions of unseen wordpieces.  Overall, our findings are promising for low-resource language varieties, demonstrating that large improvements in performance are possible with the help of a little unlabeled data, and that the performance discrepancy of multilingual models for low-resource languages \citep{wu-dredze-2020-languages} can be overcome.

%% file: sections/5_relatedwork.tex
\section{Further Related Work}\label{sec:related}

Our work builds on prior empirical studies on multilingual models, which probe the behavior and components of existing models to explain \emph{why} they are effective. \citet{Cao2020Multilingual}, \citet{pires-etal-2019-multilingual}, and \citet{wu-dredze-2019-beto} note the importance of both vocabulary overlap and the relationship between languages in determining the effectiveness of multilingual models, but they primarily consider high-resource languages.  On the other hand, \citet{wu2019emerging} and \citet{wang2019cross} find vocabulary overlap to be less significant of a factor, instead attributing such models' successes to typological similarity and parameter sharing.  \citet{artetxe-etal-2020-cross} emphasize the importance of sufficiently representing the target language in the vocabulary.  Unlike these studies, we primarily consider \emph{how} to improve the performance of multilingual models for a given target language variety.  Though our experiments do not directly probe the impact of vocabulary overlap, we contribute further evaluation of the importance of improved modeling of the target variety.

Recent work has also proposed additional pretraining for general-purpose language models, especially with respect to domain \citep{alsentzer-etal-2019-publicly,chakrabarty-etal-2019-imho,dontstoppretraining2020,han-eisenstein-2019-unsupervised,howard-ruder-2018-universal,logeswaran-etal-2019-zero,sun2019finetune}.  \citet{lakew2018transfer} and \citet{zoph-etal-2016-transfer} perform additional training on parallel data to adapt bilingual translation models to unseen target languages, while \citet{mueller-etal-2020-sources} improve a polyglot task-specific model by finetuning on labeled monolingual data in the target variety.  To the best of our knowledge, our work is the first to demonstrate the effectiveness of additional pretraining for \emph{massively multilingual} language models toward a target low-resource language variety, using only unlabeled data in the target variety.

%% file: sections/6_conclusion.tex
\section{Conclusion}

We explore additional language-specific pretraining and vocabulary augmentation for multilingual contextual word representations in low-resource settings and find them to be effective for dependency parsing, especially in the lowest-resource cases.  Our results demonstrate the significance of the relationship between a multilingual model's pretraining data and a target language.  We expect that our findings can benefit practitioners in low-resource settings, and our data, code, and models are publicly available to accelerate further study.

%% file: sections/a_acknowledgments.tex
\section*{Acknowledgments}

We thank Jungo Kasai, Phoebe Mulcaire, members of UW NLP, and the anonymous reviewers for their helpful comments on preliminary versions of this paper.
We also thank Hongmin Wang for providing the unlabeled Singlish data.
This work was supported by a NSF Graduate Research Fellowship to LHL and by NSF grant 1813153.

%% file: sections/a_appendix.tex
\section{Supplementary Material to Accompany \textit{\papertitle{}}}

This supplement contains further details about the experiments presented in the main paper.

\subsection{Vocabulary Augmentation and Statistics}\label{app:vocab}

\begin{table}[h]
    \centering
    \begin{tabular}{lrr}
        \hline
        \textbf{Language} & \textbf{Original} & \textbf{Augmented} \\
        \hline
        \ga{} & 12807 & 228 \\
        \mt{} & 49791 & 1124 \\
        \sing{} & 3 & 0 \\
        \vi{} & 6955 & 421 \\
        \hline
    \end{tabular}
    \caption{Number of tokens with unknown wordpieces in the unlabeled dataset under original and augmented vocabularies.}
    \label{tab:vocab}
\end{table}

We choose the vocabulary size to minimize the number of unknown wordpieces while maintaining a similar wordpiece-per-token ratio as the original \mbert{} vocabulary.  Empirically, we find a vocabulary size of 5000 to best meet these criteria.  Then, we tokenize the unlabeled data using both the new and original vocabularies.  We compare the tokenizations of each word and note cases where the new vocabulary yields a tokenization with fewer unknown wordpieces than the original one.  We select the 99 most common wordpieces that occur in these cases and use them to fill the 99 unused slots in \mbert{}'s vocabulary.  For Singlish, 99 such wordpieces are not available; we fill the remaining slots with the most common wordpieces in the new vocabulary.

Tab.~\ref{tab:vocab} gives a comparison of the number of tokens with unknown wordpieces under the original and augmented \mbert{} vocabularies.  The augmented vocabulary significantly decreases the number of unknowns, resulting in a specific embedding for most of the wordpieces.

\subsection{Data Extraction and Preprocessing}\label{app:preprocessing}

In this section, we detail the steps used to obtain the pretraining data.  After dataset-specific preprocessing, all datasets are tokenized with the multilingual spaCy tokenizer.\footnote{\url{https://spacy.io/models/xx}}  We then generate pretraining shards in a format acceptable by \mbert{} using scripts provided by \citet{devlin-etal-2019-bert} and the parameters listed in Tab.~\ref{tab:script_params}, which includes artificially augmenting each dataset five times by masking different words with a probability of 0.15.  Statistics for labeled datasets, which we use without modification, are provided in Tab. \ref{tab:data_labeled}.

\begin{table}[t]
    \centering
    \begin{tabular}{llrr}
        \hline
        \textbf{Language} & \textbf{Partition} & \textbf{\# Sentences} & \textbf{\# Tokens} \\
        \hline
        \multirow{3}{*}{\ga{}} & Train & 858 & 20k \\
        & Valid. & 451 & 9.8k \\
        & Test & 454 & 10k \\
        \hline
        \multirow{3}{*}{\mt{}} & Train & 1123 & 23k \\
        & Valid. & 433 & 11k \\
        & Test & 518 & 10k \\
        \hline
        \multirow{3}{*}{\sing{}} & Train & 2465 & 22k \\
        & Valid. & 286 & 2.5k \\
        & Test & 299 & 2.7k \\
        \hline
        \multirow{3}{*}{\vi{}} & Train & 1400 & 24k \\
        & Valid. & 800 & 13k \\
        & Test & 800 & 14k \\
        \hline
    \end{tabular}
    \caption{Statistics for labeled Universal Dependencies datasets.}
    \label{tab:data_labeled}
\end{table}

\paragraph{Wikipedia Data}

We draw data from the newest available Wikipedia dump\footnote{\url{https://dumps.wikimedia.org/}} for the language at the time it was obtained: October 20, 2019 (Irish) and January 1, 2020 (Maltese, Vietnamese).  We use WikiExtractor\footnote{\url{https://github.com/attardi/wikiextractor}} to extract the article text, split sentences at periods, and remove the following items:
\begin{itemize}
    \item Document start and end line
    \item Article titles and section headers
    \item Categories
    \item HTML content (e.g., \texttt{<br>})
\end{itemize}

Articles are kept contiguous.  The full Vietnamese Wikipedia consists of nearly 6.5 million sentences (141 million tokens); to simulate a truly low-resource setting, we randomly select 5\% of the articles without replacement to use in our pretraining.

\paragraph{Singlish Data}

Beginning with the raw crawled sentences from \citet{wang-etal-2017-universal}, we remove any sentences that appear verbatim in the validation or test sets of either their original treebank or our partition.  Furthermore, we remove any sentences with fewer than five tokens or more than 50 tokens, as we observe that a large proportion of these sentences are either nonsensical or extended quotes from Standard English.  We note that this dataset is non-contiguous: most sentences do not appear in a larger context.

\subsection{Training Procedure}\label{app:training}

During pretraining, we use the original implementation of \citet{devlin-etal-2019-bert} but modify it to optimize based only on the masked language modeling (MLM) loss.  Although \citet{devlin-etal-2019-bert} also trained on a next sentence prediction (NSP) loss, subsequent work has found joint optimization of NSP and MLM to be less effective than MLM alone \citep{wang2019cross,lample2019cross,liu2019roberta}.  Furthermore, in certain low-resource language varieties, fully contiguous data may not be available, rendering the NSP task ill-posed.  We perform additional pretraining for up to 20 epochs, selecting our final model based on average validation LAS downstream.

Following prior work on parsing with \mbert{} \citep{kondratyuk-straka-2019-75}, parsers are trained with a inverse square root learning rate decay and linear warmup, and gradual unfreezing and discriminative finetuning of the layers.  These models are trained for up to 200 epochs with early stopping based on the validation performance.  All parsers are implemented in AllenNLP, version 0.9.0 \citep{gardner-etal-2018-allennlp}.

Tab.~\ref{tab:script_params} gives all hyperparameters kept constant during \mbert{} pretraining and parser training.  The values for these hyperparameters largely reflect the defaults or recommendations specified in the implementations we used.  For instance, the base learning rate for \pt{}, \va{}, and \tva{} reflect recommendations in the code of \citet{devlin-etal-2019-bert}, and the \tva{} embedding learning rate is equal to the learning rate used in the original pretraining of \mbert{}.

Due to the large number of parameters in \mbert{}, large batch sizes are sometimes infeasible.  We reduce the batch size until training is able to complete succesfully on our GPU.

\elmotrained{} models are trained with the original implementation and default hyperparameter settings of \citet{peters-etal-2018-deep}.  However, following the implementation of \citet{mulcaire-etal-2019-polyglot}, we use a variable-length character vocabulary instead of a fixed-sized one to fully model the distribution in each language.  \fttrained{} is trained using the skipgram model for five epochs, with the default hyperparameters of \citet{bojanowski-etal-2017-enriching}.  All experiments are variously conducted on a single NVIDIA Titan X or Titan XP GPU.

\subsection{Hyperparameter Optimization}\label{app:hyperparam}

\begin{table}
    \centering
    \begin{tabular}{lrrrr}
        \hline
        \textbf{Representations} & \textbf{\ga{}} & \textbf{\mt{}} & \textbf{\sing{}} & \textbf{\vi{}} \\
        \hline
        \elmotrained{} & 10 & 10 & 5 & 10 \\
        \hline
        \pt{} & 5 & 20 & 5 & 5 \\
        \va{} & 10 & 15 & 1 & 5 \\
        \tva{} & 10 & 5 & 20 & 5 \\
        \hline
        \pt{} + \ft{} & 20 & 10 & 1 & 5 \\
        \va{} + \ft{} & 10 & 10 & 1 & 5 \\
        \tva{} + \ft{} & 15 & 5 & 1 & 5 \\
        \hline
    \end{tabular}
    \caption{Number of pretraining epochs used in final models, selected based on validation LAS scores.}
    \label{tab:epoch}
\end{table}

For our experiments, we fix both the pretraining and downstream architectures and tune only the number of pretraining epochs.  For \pt{}, \va{}, and \tva{}, we pretrain for an additional \{1, 5, 10, 15, 20\} epochs.  For \elmotrained{}, we pretrain for \{1, 3, 5, 10\} epochs.  Final selections are given in Tab.~\ref{tab:epoch}.

\paragraph{Measuring Variation}

We use Allentune \citep{dodge-etal-2019-show} to compute standard deviations for our experiments.  For a given representation source, we randomly select five assignments of the following training hyperparameters via uniform sampling from the ranges specified in Tab.~\ref{tab:hyperparam}.  To choose the best epoch for each method, we compute the average validation LAS for these five assignments to choose our final model.  Then, we compute the average and standard deviation of the test LAS from each of these assignments.

In cases where a hyperparameter assignment yields exploding gradients and/or trends toward an infinite loss, we rerun the experiment to yield a feasible initialization.

\begin{table}[t]
    \centering
    \begin{tabular}{lrr}
        \hline
        \textbf{Hyperparameter} & \textbf{Min} & \textbf{Max} \\
        \hline
        Adam, Beta 1 & 0.9 & 0.9999 \\
        Adam, Beta 2 & 0.9 & 0.9999 \\
        Gradient Norm & 1.0 & 10.0 \\
        Random Seed, Python & 0 & 100000 \\
        Random Seed, Numpy & 0 & 100000 \\
        Random Seed, PyTorch & 0 & 100000 \\
        \hline
    \end{tabular}
    \caption{Hyperparameter bounds for measuring variation.}
    \label{tab:hyperparam}
\end{table}

\begin{table}
    \centering
    \begin{tabular}{lrrrr}
        \hline
        \textbf{Representations} & \textbf{\ga{}} & \textbf{\mt{}} & \textbf{\sing{}} & \textbf{\vi{}} \\
        \hline
        \tva{} & 15 & 20 & 20 & 5 \\
        \tva{} + \ft{} & 15 & 15 & 5 & 5 \\
        \hline
    \end{tabular}
    \caption{Number of pretraining epochs used in original \tva{} models.}
    \label{tab:epoch_original}
\end{table}

\begin{table*}[p]
    \centering
    \begin{tabular}{llr}
        \hline
        \textbf{Stage} & \textbf{Hyperparameter} & \textbf{Value} \\
        \hline
        \multirow{4}{*}{Data Creation} & Max Sequence Length & 128 \\
        & Max Predictions per Sequence & 20 \\
        & Masked LM Probability & 0.15 \\
        & Duplication Factor & 5 \\
        \hline
        \multirow{7}{*}{Pretraining} & Max Sequence Length & 128 \\
        & Warmup Steps & 1000 \\
        & Batch Size & \{12, 16\} \\
        & Max Predictions per Sequence & 20 \\
        & Masked LM Probability & 0.15 \\
        & Learning Rate & 0.00002 \\
        & \tva{} Embedding Learning Rate & 0.0001 \\
        \hline
        \multirow{15}{*}{Parser} & Dependency Arc Dimension & 100 \\
        & Dependency Tag Dimension & 100 \\
        & \mbert{} Layer Dropout & 0.1 \\
        & \elmotrained{} Dropout & 0.5 \\
        & Input Dropout & 0.3 \\
        & Parser Dropout & 0.3 \\
        & Optimizer & Adam \\
        & Parser Learning Rate & 0.001 \\
        & \mbert{} Learning Rate & 0.00005 \\
        & Learning Rate Warmup Epochs & 1 \\
        & Epochs & 200 \\
        & Early Stopping (Patience) & 20 \\
        & Batch Size & \{8, 24, 64\} \\
        & BiLSTM Layers & 3 \\
        & BiLSTM Hidden Size & 400 \\
        \hline
    \end{tabular}
    \caption{Hyperparameters for data creation, pretraining, and parser.}
    \label{tab:script_params}
\end{table*}

\subsection{\tva{} Revision}

In April 2022, GitHub user \texttt{thnkinbtfly} reported that the original implementation of \tva{} contained an error that caused it to be equivalent to that of \va{}.  The main paper has been updated to report corrected results and best epochs for all \tva{} configurations.  Original results for all other configurations are unchanged.  We preserve the full set of original results in Tab.~\ref{tab:dependency_parsing_orig} and the original best epochs for \tva{} configurations in Tab.~\ref{tab:epoch_original}.

The original experimental conclusions about additional language-specific pretraining and vocabulary augmentation at large are unaffected: both methods are still effective as described.  Similarly, observations about the correlation between relative error reduction and the target language variety's relationship to \mbert{}'s pretraining data are unchanged.

One finding on the details of vocabulary augmentation is affected: rather than slightly benefiting learning of the newly added wordpieces, \tva{} does not consistently improve over \va{}.  Our observations in \S \ref{sec:method} and \S \ref{sec:results} have been updated accordingly.

\begin{table*}
    \centering
    \begin{tabular}{lrrrr}
        \hline
         \multirow{2}{*}{\textbf{Representations}} & \textbf{Irish (\ga{})} & \textbf{Maltese (\mt{})} & \textbf{Singlish (\sing{})} & \textbf{Vietnamese (\vi{})} \\
        & Type 1 & Type 2 & Type 0 & Type 0 \\
        \hline
        \fttrained{} & 65.36 \err{1.33} & 68.23 \err{0.61} & 66.42 \err{0.92} & 53.37 \err{0.95} \\
        \elmotrained{} & 68.25 \err{0.37} & 74.33 \err{0.53} & 68.63 \err{1.04} & 56.91 \err{0.41} \\
        \hline
        \mbert{} & 68.19 \err{0.43} & 67.06 \err{0.61} & 74.01 \err{0.39} & 62.96 \err{0.41} \\
        \pt{} & \textbf{73.03 \err{0.25}} & 78.51 \err{0.41} & \textbf{76.48 \err{0.63}} & \textbf{64.67 \err{0.22}} \\
        \va{} & 72.68 \err{0.47} & \textbf{79.88 \err{0.55}} & \textbf{76.71 \err{0.70}} & 64.28 \err{0.44} \\
        \tva{} & \textbf{73.11 \err{0.37}} & 79.32 \err{0.45} & \textbf{76.92 \err{0.77}} & \textbf{64.46 \err{0.44}} \\
        \hline
        \mbert{} + \ft{} & 72.67 \err{0.22} & 76.74 \err{0.35} & 78.24 \err{0.52} & 66.13 \err{0.38} \\
        \pt{} + \ft{} & 75.45 \err{0.28} & 82.77 \err{0.24} & 79.30 \err{0.57} & \textbf{67.50 \err{0.25}} \\
        \va{} + \ft{} & \textbf{76.17 \err{0.08}} & \textbf{83.53 \err{0.21}} & \textbf{79.89 \err{0.46}} & 67.28 \err{0.38} \\
        \tva{} + \ft{} & \textbf{76.23 \err{0.22}} & 83.16 \err{0.25} & \textbf{80.09 \err{0.34}} & \textbf{67.82 \err{0.27}} \\
        \hline
    \end{tabular}
    \caption{Original results (LAS) on downstream UD parsing.
    }
    \label{tab:dependency_parsing_orig}
\end{table*}